\documentclass[conference]{IEEEtran}
\IEEEoverridecommandlockouts
\usepackage{cite}
\usepackage{amsmath,amssymb,amsfonts}
\usepackage{algorithmic}
\usepackage{graphicx}
\usepackage{textcomp}
\usepackage{xcolor}
\usepackage{booktabs}
\usepackage{bm}
\usepackage[caption=false]{subfig}

\def \R{\mathbb{R}}
\def \W{\mathbf{W}}
\def \ks{\mathbf{k}}
\def \vs{\mathbf{v}}
\def \q{\mathbf{q}}
\def \h{\mathbf{h}}
\def \KB{\mathbf{K}}
\def \w{\mathbf{w}}
\def \x{\mathbf{x}}

\DeclareUnicodeCharacter{2212}{-}

\DeclareMathOperator*{\argmin}{arg\,min}
\newcommand{\CE}{\operatorname{CrossEntropy}}
\usepackage[breaklinks,colorlinks]{hyperref}
\def\BibTeX{{\rm B\kern-.05em{\sc i\kern-.025em b}\kern-.08em
    T\kern-.1667em\lower.7ex\hbox{E}\kern-.125emX}}
\begin{document}

\title{Deep Dynamic Effective Connectivity Estimation from Multivariate Time Series}


\author{\IEEEauthorblockN{Usman Mahmood $^{1,2}$ \IEEEauthorrefmark{1},
Zening Fu$^{1,2}$, Vince Calhoun$^{1,2,3,4}$ and
Sergey Plis$^{1,2}$}
\IEEEauthorblockA{$^{1}$ Tri-institutional Center for Translational Research in Neuroimaging and Data Science (TReNDS),\\ $^{2}$ Georgia State University, $^{3}$ Georgia Institute of Technology, $^{4}$Emory University\\
Correspondence: \IEEEauthorrefmark{1}umahmood1@gsu.edu.
}}



\maketitle

\begin{abstract}
Recently, methods that represent data as a graph, such as graph neural networks (GNNs) have been successfully used to learn data representations and structures to solve classification and link prediction problems.
The applications of such methods are vast and diverse, but most of the current work relies on the assumption of a static graph.
This assumption does not hold for many highly dynamic systems, where the underlying connectivity structure is non-stationary and is mostly unobserved.
Using a static model in these situations may result in sub-optimal performance.
In contrast, modeling changes in graph structure with time can provide information about the system whose applications go beyond classification.
Most work of this type does not learn effective connectivity and focuses on cross-correlation between nodes to generate undirected graphs.
An undirected graph is unable to capture direction of an interaction which is vital in many fields, including neuroscience.
To bridge this gap, we developed dynamic effective connectivity estimation via neural network training (DECENNT), a novel model to learn an interpretable directed and dynamic graph induced by the downstream classification/prediction task.
DECENNT outperforms state-of-the-art (SOTA) methods on five different tasks and infers interpretable task-specific dynamic graphs.
The dynamic graphs inferred from functional neuroimaging data align well with the existing literature and provide additional information.
Additionally, the temporal attention module of DECENNT identifies time-intervals crucial for predictive downstream task from multivariate time series data.
\end{abstract}


\section{Introduction}
Many classification/prediction problems can be solved by learning the underlying structure/pattern of the data and how different components are co-related with each other. Datasets from different fields are often represented as a graph. Graph networks \cite{4700287, bruna2014spectral} are proposed to work on such datasets. Recently, methods such as, graph neural networks (GNNs) have been extensively used to learn representations on graph-structured data \cite{Bronstein_2017,hamilton2018representation, gilmer2017neural, PARISOT2018117}. GNNs take nodes from data and update representations of nodes with the help of different aggregating functions. The aggregate functions work using a message-passing system, where a node receives messages from its neighbors, which are defined by edges. The representations can then be used for node classification, graph classification, or predicting edges between nodes by using an existing true graph structure or learning the graph \cite{monti2016geometric,velickovic2018graph,kipf2017semisupervised, gilmer2017neural, 10.1093/bioinformatics/bty294,zhang2018link, wang2019dynamic, kipf2018neural,Zitnik_2018}. For any of the mentioned tasks, most of the existing work (classification, link prediction) has been done on static graphs, e.g., Parisot~et~al.~\cite{PARISOT2018117}, creates a static graph based on representation and phenotype information of subjects, Mahmood~et~al.~\cite{Mahmood_2021} learns a static graph between brain regions, Kipf~et~al.~\cite{kipf2018neural} learns a static graph in an interacting system. In reality, many fields (social networks, brain connectivity, traffic data, speech) are dynamically changing and cannot be completely represented using a static graph. We propose that learning a dynamic graph for such systems can increase our understanding of these highly dynamic systems and may also yield higher classification performance based on the task. For example, learning the dynamic connectivity of the brain's networks can help researchers to understand brain dynamics and the causes of brain disorders by learning how connectivity changes while performing tasks, or with age. Dynamic graphs for social network data can help to understand users' patterns, peak traffic times, retention time, and many other vital aspects of the network.

Even though graph networks have excelled in many areas, we see a couple of shortcomings in the current work regarding graph-structured data. 1) Most of the work done, whether for graph classification (node or graph) or graph learning (link prediction), work on un-directed graphs and assume that the graph structure of the data is available or easily created and thus work directly on the graph-structured data \cite{li2017gated, gilmer2017neural}. This assumption is improbable in many different datasets across many fields. 2) The semi or unsupervised methods developed to learn the graph structure (link-prediction) create embeddings/representations based on the learned graph structure, and use these embeddings to either predict future embeddings or perform classification tasks where loss is the error in prediction or classification \cite{kipf2018neural, kazi2020differentiable, Mahmood_2021}. The problem with this approach is that the embeddings are used for prediction/classification. The learned structure is not tested directly, especially where the true graph structure is never available (e.g., brain functional network connectivity (FNC)) and thus is unreliable. The unreliability increases in systems with relatively easy tasks and noiseless data (real or simulated). Kipf~et~al.~\cite{kipf2018neural} shows that using a full graph leads to almost the same or better performance (in terms of loss) as the learned static graph, thus questioning the correctness/importance of the learned graph structure. 3)~The graph structure is assumed to be static, which is highly unlikely in many datasets such as a) social network, where a node can join/leave at any time or create/drop an edge. b) Brain functional connectivity, where the connectivity between brain regions is always dynamic. Xu~et~al.~\cite{Xu2020Inductive} shows that using a static graph learning method for a dynamic system/graph can lead to lower classification performance, Kipf~et~al.~\cite{kipf2018neural} shows improved results by just dynamically re-evaluating the static learned graph during testing. The improved performance for the relevant task is understandable as the dynamic connectivity provides essential information about the system. 

Classification gains by using a dynamic graph depend on the downstream task; e.g., a social network's dynamic graph may not be too helpful to predict a user's gender but can provide additional information, which itself is extremely important for understanding the dynamic system and its working. For example, studies like \cite{article21,article22,CALHOUN2014262} show that dynamic functional connectivity (FC) show re-occurring patterns which cannot be captured in static FC. Thus, we want to learn dynamic and directed graph structure representing the time series data and use that for interpretation, understanding and prediction tasks.

We present a novel method called - dynamic effective connectivity estimation via neural network training - (DECENNT). DECENNT is a semi-supervised method (unknown graph structure, known graph labels) which we use to 1) learn dynamic directed graph structure using embeddings for better understanding of the underlying system and 2) perform graph classification based on the learned graph structure/connectivity alone. We propose that high classification results based on the learned graph structure alone and not the representations greatly reduce the uncertainty regarding the usefulness of the learned structure and produce more useful and reliable graphs, which is the key objective of our study. These graphs can then be used for understanding the underlying system and interpreting the cause(s) of classification/prediction.

Recently, graph structure has been used to represent the brain. Brain connectivity is highly dynamic and changes with functionality being performed. Understanding the dynamic functionality would help to understand the functionality and connectivity of the brain. Thus, we apply our model to learn the brain's dynamic effective connectivity (EC) using the functional magnetic resonance imaging (fMRI) data. fMRI is an imaging method used to capture blood-oxygenation level dependent (BOLD) signals in the brain, which measures neural activity between brain regions. 

In this study, we learn dynamic EC of the brain and use that to 1) predict the presence of a disease or predict the gender of the subject and 2) learn dynamics of brain networks' connectivity related to the downstream task. Many recent studies have been proposed to learn or compute FC of the brain \cite{arslan2018graph,10.3389/fnins.2020.00630,ktena2017distance,ma2019similarity,KTENA2018431} and use it to predict the gender or disease/disorder \cite{mahmood2021brain,arslan2018graph,kazi2021iagcn,10.3389/fnins.2020.00630,KTENA2018431,ma2019similarity,mahmood2021multi} using GNNs or other such methods but have limitations as discussed above. Methods that incorporate dynamic FC (either learned or computed) are mostly window-based \cite{DAMARAJU2014298, ARMSTRONG2016175,gadgil2021spatiotemporal,10.1007/978-3-030-59861-7_1}, which partition the data into multiple windows, each consisting of data from multiple time-points. As the structure/connectivity can change at any time, we create an instantaneous dynamic structure. Existing studies most commonly use symmetric estimates of relationship, e.g., Pearson correlation coefficients (PCC) matrix, to represent FC. The symmetric correlation matrix does not capture effective connectivity, and does not reveal the direction of flow of information. To incorporate direction, some studies \cite{EntropyApp,grangerApp} use methods like transfer entropy \cite{entropybook} or Granger causality \cite{grangercausality}. The former is notoriously difficult to estimate reliably, while the latter has problems with latent confounders. Moreover, these approaches do not directly fit the requirements for an interpretable differential layer. 

We also incorporate temporal attention in our model to provide better interpretable results and further understand the working of brain functionality. We apply our model to both resting state-fMRI (rs-fMRI), where the essential time-points (putative biomarkers) are not known and to speech data. In the latter, we predict the presence of a specific target word in the speech and use the attention module to mark the time-points of the occurrence.

\textbf{Contributions:}
Our study has the following contributions.
\begin{enumerate}
    \item Without the availability brain's true E/FC structure, by using fMRI data, we learn a directed connectivity structure of the brain that provides additional details than existing literature. Thus, it removes the need to use a separate method to compute connectivity before applying classification.
    \item Based on the learned dynamic EC, our model outperforms other SOTA methods in classification tasks (disorders, gender, and keyword detection) across multiple datasets and pre-processing.
    \item Our temporal attention module finds the essential time-points for the downstream task with very high accuracy and is stable/consistent across multiple trials. It improves classification performance and finds important bio-markers related to the downstream task. This in turn can lead to reducing the temporal dimensions and discarding time-points that are unrelated to the downstream task.
\end{enumerate}

\label{DECENNT}
\section{DECENNT}

\begin{figure}[t]
\centering
\includegraphics[width=\columnwidth]{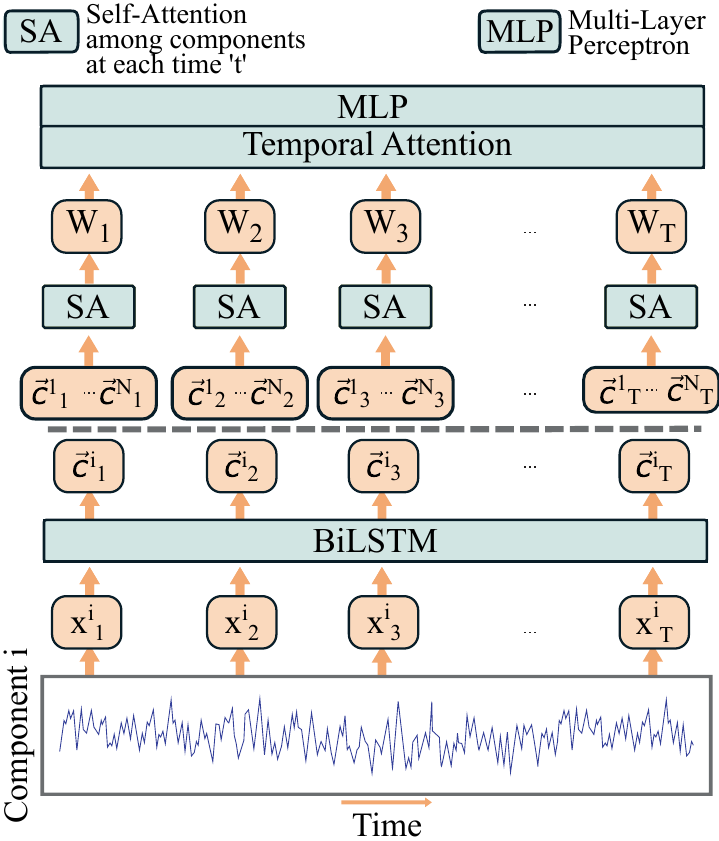}
\caption{DECENNT architecture using biLSTM, self-attention and temporal attention.}
\label{fig:TSA - Arch}
\end{figure}

We use the proposed DECENNT model to learn a dynamic directed graph structure/pattern for any multivariate time series data. The dynamic directed graph is essential to learning and understanding the system and can be used in different ways to perform classification on the downstream task. We learn a distinct dynamic graph $G$ for the complete time series where $G$ is a set of $T$ graphs with $T$ being the total time-points of the time series. We define G as: 

$G =  \{ g_1, .... g_t, .... g_T | T =$ time-points\} and $g_t = (V_t, E_t)$, where, $V_t$ and $E_t$ represent the vertices and edges present at time-point $t$.  After computing the set $G$ we use our temporal attention module to focus on the important time-points and generate a single final graph $G_f$ representing the complete time series. $G_f$ is used for downstream classification. To create the embedding $h_{i,t}$ for the $i_{th}$ component at time $t$ we use a bidirectional long short-term memory (biLSTM) \cite{650093} which takes the time series for the component $i$ and produces $h_t$ for each component. To create the connectivity matrix (adjacency matrix/graph) between the components (nodes) at each time-point $t$ we use a self-attention module \cite{10.5555/3295222.3295349}. We explain both parts separately in the following sections. Fig.~\ref{fig:TSA - Arch} shows the complete architecture of the model.

\subsection{biLSTM}

biLSTMs have been used very successfully for time series data. LSTMs take one input (e.g., word) from a sequence (e.g., sentence) and provide embeddings for data at each point in the sequence. The effectiveness of LSTMs comes from the memory and forget gates, which help the model to learn relationships between input at different time-points. In time series data, e.g., a sentence, each input is not independent of previous or future values. Thus it makes very crucial to find these effective relationships between the data. As the effect of input at a time, $t$ onto other inputs, is unknown and can vary across different time series and components of the same time series, it is crucial to learn these relationships based on the downstream task. The working of the LSTMs can be explained by the following equations. $\sigma$ represents sigmoid activation, $b$ are the biases, and $\odot$ is the Hadamard product \cite{Million07thehadamard}. 
\begin{equation}
\begin{array}{ll}

            \mathbf{i_t} = \sigma(\W_{ii} \x_t + b_{ii} + \W_{hi} \h_{t-1} + b_{hi}) \\
            \mathbf{f_t} = \sigma(\W_{if} \x_t + b_{if} + \W_{hf} \h_{t-1} + b_{hf}) \\
            \mathbf{g_t} = \tanh(\W_{ig} \x_t + b_{ig} + \W_{hg} \h_{t-1} + b_{hg}) \\
            \mathbf{o_t} = \sigma(\W_{io} \x_t + b_{io} + \W_{ho} \h_{t-1} + b_{ho}) \\
            \mathbf{c_t} = \mathbf{f_t} \odot \mathbf{c_{t-1}} + \mathbf{i_t} \odot \mathbf{g_t} \\
            \h_t = \mathbf{o_t} \odot \tanh(\mathbf{c_t}) \\
\end{array}
\end{equation}
Here $\h_t$ represent the embedding for the input at $t$. We use a biLSTM to create embeddings $\h_t$ for each component $i$. Thus $\h_t^f = LSTM(x_t,\h_{t-1})$, $\h_t^b = LSTM(x_t,\h_{t+1})$ and $\h_t = concatenate(\h_t^f,\h_t^b)$. Here $\h_t^f$ and $\h_t^b$ are representation for forward and backward pass. We use LSTM for each component individually, sharing weights of LSTM among the components. We give $x_t^i$ as input to the LSTM along with hidden vector and receive $\h_t^i$ for the component $i$. This allows us to later compute connectivity matrix (links/edges) between the components/nodes.

\section{Self-Attention}

Self-attention creates new embeddings for each $\mathbf{x^i}$ depending on $n$ other embeddings in the sequence. Self-attention tries to find the relationship of each input with all other inputs denoted by weights and updates $\mathbf{x^i}$ accordingly. Self-attention can be considered a special case of a typical GNN with $1$ layer/hop where a node receives input from the neighbors that are one hop away. Because of the ability of the self-attention module to create weights by learning the relationship between different embeddings, we create a connectivity matrix between components at each time-point $t$ by giving \{$\h{}_t^i,..... \h{}_t^n$ \}, $n$ = total components, as input to the self-attention module and creating new embeddings \{$\tilde{\h{}}_t^i,..... \tilde{\h{}}_t^n $\} and the weight matrix $\W{}_t$, where each $\W{}_t \in \R{}^{n*n}$. The self-attention module creates three embeddings, namely, key ($\ks{}$), value ($\vs{}$), and query ($\q{}$) and creates new embeddings for each input using these embeddings. The set of equations in \eqref{Eq: SA} can sum up the whole process. For simplicity, we omit the $t$ from these equations. $^\intercal$ represents transpose and $\oplus$ represents concatenation.   

\begin{equation}
\begin{array}{ll}
{\ks{}^i} = {{\h{}^i}^\intercal \W{}^{(k)}} , \hspace{.25cm}  {\vs{}^i} = {{\h{}^i}^\intercal \W{}^{(v)}},\hspace{.25cm}
{\q{}^i} = {{\h{}^i}^\intercal \W{}^{(q)}}\\
{\KB{}} = \oplus_{i = 1}^n {\ks{}^i}^\intercal, \hspace{.25cm}  
{\w{}^i} = \operatorname{softmax}({\q{}^i} {\KB{}})\\
{\tilde{\h{}}^i} = \sum_j^n{({w^i_j} {\vs{}^j})}, \W = \oplus_{i = 1}^n \w{}^i
\end{array}
\label{Eq: SA}
\end{equation}
Here $\W{} \in \R{}^{n*n}$ is the connectivity matrix between $n$ components/nodes in the graph. We use $\W{}$ for downstream classification and not the embeddings. As the true graph is never available in many applications to directly compare with, we propose that a connectivity matrix leading to state-of-the-art classification performance makes it more reliable than using the embeddings ($\tilde{\h{}}$) for classification.

\section{Temporal Attention}

Since we get a set of the matrices $\W{}$, one for each time-point, an easy and standard way is to average the $T$ matrices, however, not all time-points can be equally crucial for the downstream task. Therefore, we introduce a new temporal attention mechanism to focus on important time-points. The temporal attention module is essential for the downstream classification and for finding important timepoints/biomarkers in the data, thus making it crucial to our model. We name our attention model - global temporal attention (GTA) - that attends to crucial time-points and is stable and consistent across randomly seeded trials.

\subsection{GTA}
\label{SectionGTA}
To give the attention module a global view of the graph, we present GTA. The global view allows the model to learn how each connectivity matrix contributes to the global graph or structure of the data in the downstream task. We create a sum of all the $T$ connectivity matrices and call it $\W{}^{global}$ representing the global view. We then compare the similarity of each local $\W_t$ with the global view and use them to create the temporal attention vector $\bm{\alpha}$.

\begin{equation}
\begin{array}{ll}
\W{}^{global} = \sum_{t = 1}^T (\W{}_t), 
\widetilde{\W{}_t} = \W{}_t \odot \W{}^{global}\\
\bm{\alpha} =  (\oplus_{t = 1}^T ( ((\operatorname{flat} (\widetilde{\W{}_t})) \W{}^{MLP_{l1}}) \W{}^{MLP_{l2}})\\
\W{}^f = \sum_{t = 1}^T (\W{}_t {\alpha_t} )
\end{array}
\label{eq:GTA}
\end{equation}

Here $\odot$ is the Hadamard product \cite{Million07thehadamard} between matrices, $\oplus$ represents concatenation, and $\W^f$ is the final weight matrix. Equation \eqref{eq:GTA} shows the equations for GTA. 




\begin{table*}[ht]
\caption{Details of the neuroimaging datasets used. We tried different CV folds in our experiments but that did not have a significant effect on results. We report the results with CV folds that match comparing studies.}
\centering
\begin{tabular}{@{}lllllllll@{}}
\toprule
Name  & Category      & Preprocessing    & Parcellation & Subjects & 0 Class & 1 Class & CV Folds   & TP       \\ \midrule
FBIRN & Schizophrenia & SPM12            & ICA          & 311      & 151     & 160     & 4, 6, 18     & 157      \\
OASIS & Dementia    & SPM12            & ICA          & 912      & 651     & 261     & 4, 10          & 157      \\
ABIDE & Autism        & SPM12            & ICA          & 569 (TR=2)      & 255     & 314     & 5, 10          & 140      \\
ABIDE & Autism        & SPM12            & ICA          & 869      & 398     & 471     & 5, 10          & 140      \\
HCP   & Gender        & SPM12            & ICA          & 833  & 390     & 443     & 5, 15 & 980 \\ \cmidrule(r){1-9}  
FBIRN & Schizophrenia & SPM12            & Shaefer 200  & 311      & 151     & 160     & 18         & 157      \\
HCP   & Gender        & Glassier & Shaeffer 200 & 942      & 411       & 531       & 10         & 1200     \\
ABIDE   & Autism        & C-PAC & Shaeffer 200 & 871      & 403       & 468       & 10         & 83-316     \\
\bottomrule
\end{tabular}
\label{table:demographics}
\end{table*}

\section{Experiments}
This section reports the training process of our model, details about hyper-parameters, and datasets used.

\subsection{Training}
We ran our experiments using RTX 2080 using PyTorch. The hidden dimensions for the LSTM, self-attention including key, query, and value modules, were all set to $64$. Both LSTM and self-attention modules had only one layer. We tried to incorporate multiple layers, but it did not help in terms of classification performance nor interpretation. The dimensions of MLP layer for calculating temporal attention vector were $\gamma * len(flat(\W_t))$ and $1$ with $\gamma = 0.25$. We used batch normalization after the first MLP layer. ReLU activation was used in our model between the MLP layers. A final two-layer MLP was used to get logits for binary classification problem with $\W_f$ as input with dimensions $64$ and $2$. We used cross-entropy loss with Adam optimizer. Let $\theta$ represent all the parameters of the architecture, $\mathbf{\hat{y}}$ being the prediction and $\mathbf{y}$ is the true labels, the loss is calculated as:

\begin{gather}
loss = \CE(\mathbf{\hat{y}}, \mathbf{y} ) + \lambda \|\bm{\theta}\|_1 \\
\bm{\theta^*} = \argmin_{\bm{\theta}} (loss)
\end{gather}

$\lambda$ (regularization weight) was set to $1e^{-6}$ and learning rate ($\eta$) was $1e^{-4}$. We reduced the learning rate by a factor of $0.5$ when validation loss reached plateau. Early stopping was used to stop training the model based on validation loss and patience of $15$. For each dataset, to have a fair result, we perform n-fold cross validation, depending on the size of the data with 10 randomly seeded trials for each fold. We report the mean area under curve - receiver operating characteristic (AUC-ROC) and many other metrics to show classification performance. For region based experiments, $\gamma$ was reduced to $0.005$, $\eta$ was set to $5e-3$ for HCP dataset and $3e-4$ for others. Batch size was set to $32$.

\subsection{Datasets}
To test our model for a) classification b) learned connectivity matrix and c) learned temporal attention we use five different datasets across two fields; neuroimaging, and natural language processing (NLP). Refer to  Tab. \ref{table:demographics} for details of the neuroimaging datasets. Validation and test size was kept same.

\subsubsection{NeuroImaging}
\label{sec:neuroimaging}
The neuroimaging datasets can be further divided into two sub-tasks; brain disorder and gender prediction.

\paragraph{Disorder Prediction}
Three datasets used in this study include 
FBIRN (Function Biomedical Informatics Research Network\footnote[1]{We are using fBIRN phase III.})~\cite{keator2016function} project, release 1.0 of ABIDE (Autism Brain Imaging Data Exchange\footnote[2] {http://fcon\_1000.projects.nitrc.org/indi/abide/})~\cite{di2014autism} and release 3.0 of OASIS (Open Access Series of Imaging Studies\footnote[3]{https://www.oasis-brains.org/})~\cite{rubin1998prospective} to predict schizophrenia, autism and dementia respectively.

\paragraph{Gender Prediction}
Healthy controls from the HCP~\cite{van2013wu} are used for gender prediction. 

\paragraph{Preprocessing}
We use different brain parcellation techniques which can be divided into two sub-categories; ICA and region based. The preprocessing method applied depends on the parcellation technique used and the methods used in SOTA studies for the particular dataset.

\textit{ICA parcellation:}
\label{sec:ICA-parcellation}
All experiments used a fully automated independent component analysis (ICA) as a brain parcellation technique. We first preprocess the fMRI data using statistical parametric mapping (SPM12, http://www.fil.ion.ucl.ac.uk/spm/) within MATLAB 2020. Subjects were included in the analysis if the
subjects have head motion $\le 3^\circ$ and $\le 3$ mm, and with functional data providing near full brain successful normalization~\cite{fu2019altered}. For each subject, $100$ ICA components are estimated using the Neuromark template and used in experiments following the same procedure described in~\cite{fu2019altered}. We use ICA timecourses as input to the model. For ABIDE1, we conduct two ICA based experiments using all subjects and subjects with $TR=2$. 

\textit{Region parcellation:} SOTA methods use different preprocessing pipelines for HCP and ABIDE dataset. For comparison with these SOTA methods on HCP and ABIDE dataset, we preprocess these datasets following existing studies. HCP \cite{van2013wu} was first minimally pre-processed following \cite{article}, and then FIX-ICA based denoising was applied to reduce noise in the data \cite{article10,Griffanti2014ICAbasedAR}. After denoising, $152$ subjects were discarded based on head motion following \cite{10.3389/fnins.2020.00630} which results into $942$ subjects. ABIDE1 \cite{di2014autism} was pre-processed using cpac \cite{article9}, out of $1112$ subjects $871$ were selected following \cite{ABRAHAM2017736,PARISOT2018117,CAO2021103015}. To divide the data into regions, we use Shaefer \cite{10.1093/cercor/bhx179} and Harvard Oxford (HO) \cite{DESIKAN2006968} atlas depending on the experiment. Refer to  Tab. \ref{table:demographics} for details about the datasets. 


\subsubsection{NLP}

To show the broad implications of our method, we apply our method for keyword detection in audio files. We choose this problem because it has many practical applications (e.g., virtual assistants in smartphones and robots). 
We use Speech Commands Dataset~\cite{speechcommand} for predicting the occurrence of a keyword in speech. The audio files are combined with a background noise of a coffee shop ~\cite{backgroundnoise} to make prediction harder. We use this dataset to test the temporal attention weights we get from our model because important time-points (location of word cat in the noise) is known. We match this experiment with classifying brain disorder. The keyword "cat" can be thought as the presence of a disease and the background noise can be considered as the noise and other data present in the fMRI time-courses.

\begin{figure}[ht]
\centering
\includegraphics[width=\columnwidth]{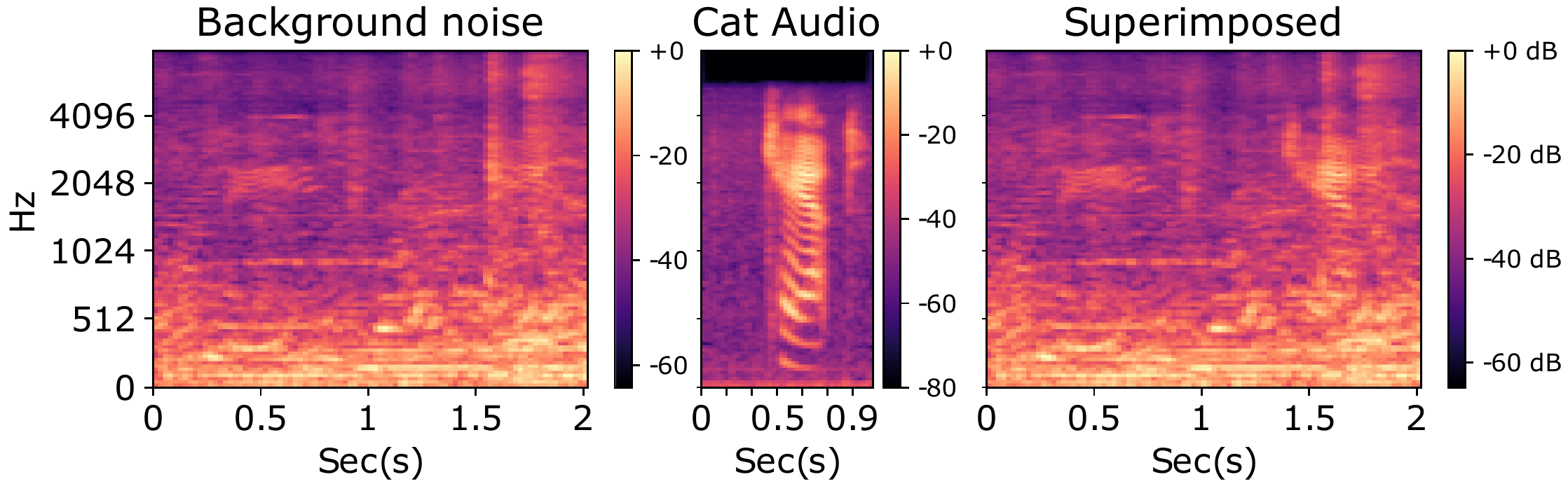}
\caption{Mel-spectogram of 'background noise', 'cat audio' and 'superimposed audio'. The model receives background and superimposed files as input.}
\label{fig:melspectogram-comparison}
\end{figure}

\paragraph{Preprocessing}
For prediction, we collect samples of audios for the keyword "cat" from speech command dataset~\cite{speechcommand} which has 1515 audio files for the keyword. To create "cat" class examples, we superimpose each of the keyword audios with the length of one second onto a two seconds long background noise at a random location, resulting in a two seconds long audio consisting of background noise and keyword cat. To make it a difficult problem, we do the following things; a) the audio of cat is mixed with the noise when creating the final audio, which means the timepoints where the word cat is added has noise as well and b) before mixing the two audio files, we match the amplitudes of the two audios by normalizing both audios (background and cat) to same scale and finally c) we normalize the 1-second long sum of both files so that the sum does not have higher values than the rest of the audio file which only has noise. As a result of the points (a-c) metioned above, the model cannot perform classification based on amplitude. Furthermore, the model receives the mel-spectogram of the audio files as input rather than actual audio files.  Fig. \ref{fig:melspectogram-comparison} shows the mel-spectogram of the three audio files. To create "no-cat" class examples, we use another 1515 two seconds long audio files containing only the background noise. Thus, we use 3030 audio files for the downstream task ("cat"/"no-cat" classification). For all of the 3030 audio files, we compute mel-spectrogram to convert each of them into a matrix of size \emph{components $\times$ time courses}. To test our model on multiple keywords, we create another dataset with keyword 'nine' following the same method.

\section{Results}

We show three different results, one for each of the paper contributions. We compare our results with SOTA DL methods  \cite{mahmood2019learnt,Mahmood_2020, Mahmood_2021,gadgil2021spatiotemporal, 10.3389/fnins.2020.00630,article7,10.1093/cercor/bhz129,arslan2018graph,CAO2021103015,PARISOT2018117,KTENA2018431} depending on the task, and ML methods such as support vector machine (SVM) and logistic regression (LR). To be fair to the other papers, we report directly from the results mentioned in the papers. Not all methods were applicable to each of the dataset/task, or the code/results of other methods were not available. All figures are generated using multiple test subjects across at least $10$ randomly seeded trials. Our experiments show that our model beats SOTA methods on classification/prediction tasks but more importantly our learned EC structures are a) similar to existing studies, b) provides knowledge not present in existing methods; FC, c) captures direction of connectivity and d) finds important temporal bio-markers relevant to the downstream task.

\begin{figure*}[t]
\centering
\includegraphics[width=\linewidth]{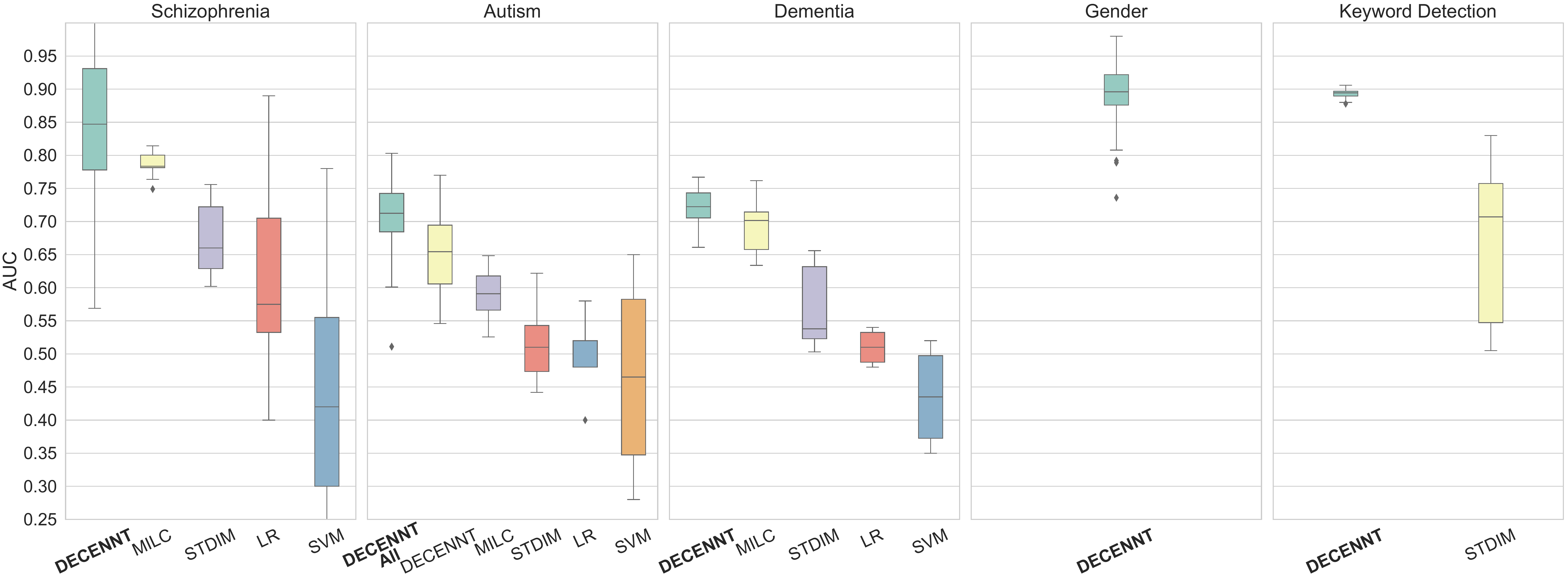}
\caption{AUC comparision of DECENNT model with four different methods (MILC \cite{Mahmood_2020}, STDIM \cite{mahmood2019learnt}, LR, SVM), over five different datasets on ICA time courses (Ref to section \ref{sec:ICA-parcellation}). Our method significantly outperforms SOTA methods. We performed Autism experiments with 869 subjects (DECENNT ALL - all TRs) as well. As we do not have a pre-training step we compare with not-pre-trained (NPT) version of MILC and STDIM. Input to ML models was same ICA time courses.}
\label{fig:AUC-Comparision}
\end{figure*}

\begin{table*}[ht]
\caption{Classification performance comparison of DECENNT with other DL methods on region based data of HCP and FBIRN datasets (Ref to section \ref{sec:ICA-parcellation}). Our DECENNT model outperforms all other methods in almost every metric. The best two scores are shown as bold and italic respectively. Note: As we use all the regions in the atlas we report the mean accuracy for SVM-RBF~\cite{10.1093/cercor/bhz129}. The results for GCN \cite{arslan2018graph} on HCP data are reported in GIN paper~\cite{10.3389/fnins.2020.00630}.}
\centering
\begin{tabular}{lllllll|ll}
\hline
\multicolumn{7}{c|}{HCP}
& \multicolumn{2}{c}{FBIRN} \\
          & DECENNT       & GIN            & SVM-RBF & GCN    & ST-GCN         & PLS  & DECENNT     & BrainGNN    \\
          \hline
AUC        & \textbf{93.6} & NA             & NA      & NA    & NA & \textit{88.125}   & \textbf{82.5}        & \textit{78.8}        \\
ACC        & \textbf{86.0} & \textit{84.6}  & 68.7    & 83.98 & 83.7          & 79.9 & NA          & NA          \\
Precision  & \textbf{87.2}   & \textit{86.19} & NA      & 84.59 & NA             & NA   & NA          & NA          \\
Recall     & \textbf{88.6} & 86.81 & NA      & \textit{87.78} & NA             & NA   & NA          & NA          \\
Parcellation &
  \begin{tabular}[c]{@{}l@{}}Shaefer \\ 200\end{tabular} &
  \begin{tabular}[c]{@{}l@{}}Shaefer\\ 400\end{tabular} &
  \begin{tabular}[c]{@{}l@{}}Shaefer 400 \\ + Fan 39\end{tabular} &
  \begin{tabular}[c]{@{}l@{}}Shaefer\\ 400\end{tabular} &
  \begin{tabular}[c]{@{}l@{}}Multi-moda\\ 22\end{tabular} &
  \begin{tabular}[c]{@{}l@{}}Dosenbach\\ 160\end{tabular} &
  \begin{tabular}[c]{@{}l@{}}Shaefer \\ 200\end{tabular} &
  \begin{tabular}[c]{@{}l@{}}AAL \\ 116\end{tabular} \\
Validation & 10            & 10             & 10      & 10    & 5             & 10   & 18          & 18          \\
Subjects   & 942           & 942            & 434     & 942   & 1091               & 820  & 311         & 311 \\ 
Study   & Our           & \cite{10.3389/fnins.2020.00630}            & \cite{10.1093/cercor/bhz129}     & \cite{arslan2018graph}   & \cite{gadgil2021spatiotemporal}               & \cite{article7}  & Our         & \cite{Mahmood_2021} \\
\bottomrule
\end{tabular}
\label{table:AUC-Comparison}
\end{table*}

\begin{table}[ht]
\caption{Comparison of AUC score on ABIDE1 region based dataset (Ref to section \ref{sec:ICA-parcellation}). Existing methods use Harvard Oxford (HO) parcellation with $111$ brain regions. Unlike GCN~\cite{PARISOT2018117} and DeepGCN~\cite{CAO2021103015} we use only fMRI data. }
\centering
\begin{tabular}{@{}llll@{}}
\toprule
Method          & Parcellation   & Input     & AUC  \\ \midrule
DECENNT         & Shaefer    & fMRI data & \textit{0.70} \\
DECENNT         & HO & fMRI data & {0.69} \\
GCN \cite{PARISOT2018117}     & HO & \begin{tabular}[c]{@{}l@{}}fMRI +\\ phenotypic data\end{tabular} & \textbf{0.75} \\
DeepGCN \cite{CAO2021103015} & HO & \begin{tabular}[c]{@{}l@{}}fMRI +\\ phenotypic data\end{tabular} & \textbf{0.75} \\
Metric Learning \cite{KTENA2018431} & HO & fMRI data & 0.58 \\ \bottomrule
\end{tabular}

\label{table:AUC-Comparison-ABIDE}
\end{table}

\subsection{Classification}

Our method turns out to be the best performing model against SOTA methods, giving the highest AUC score for all the datasets used for classification (disorder, gender, speech) with ICA data. Even with region-based data our model performs better than existing methods on HCP and FBIRN dataset. As our model does not use phenotypic information about subjects, our model lacks behind GCN ~\cite{PARISOT2018117} and DeepGCN~\cite{CAO2021103015} on ABIDE.

Parisot~et~al.~\cite{PARISOT2018117} reports a decrease of $\sim 2.5$ AUC by using a different phenotypic information which clearly shows the dependence on phenotypic data. \cite{KTENA2018431} reports much lower AUC score by using only fMRI data.   Fig. \ref{fig:AUC-Comparision} shows the classification results on ICA data. The machine learning methods fail due to high data dimensions ($m$), and relatively smaller number of subjects($n$), $m>>n$. 
Tab. \ref{table:AUC-Comparison} and Tab.  \ref{table:AUC-Comparison-ABIDE} show region based classification results. We would like to point that GIN \cite{10.3389/fnins.2020.00630} and ST-GCN \cite{gadgil2021spatiotemporal} use test data for hyper-parameter tuning and early stopping, whereas we use validation data for both and test data is used only to test the model. Kim~et~~al.~\cite{kim2021learning} reports lower results for GIN ($81.34$ ACC and $89.55$ AUC) and ST-GCN when not using test data as validation data. 

\subsection{Connectivity Matrix}

\begin{figure}[ht!]
\centering
\subfloat[Cat EC]{
\includegraphics[height=0.475\columnwidth]{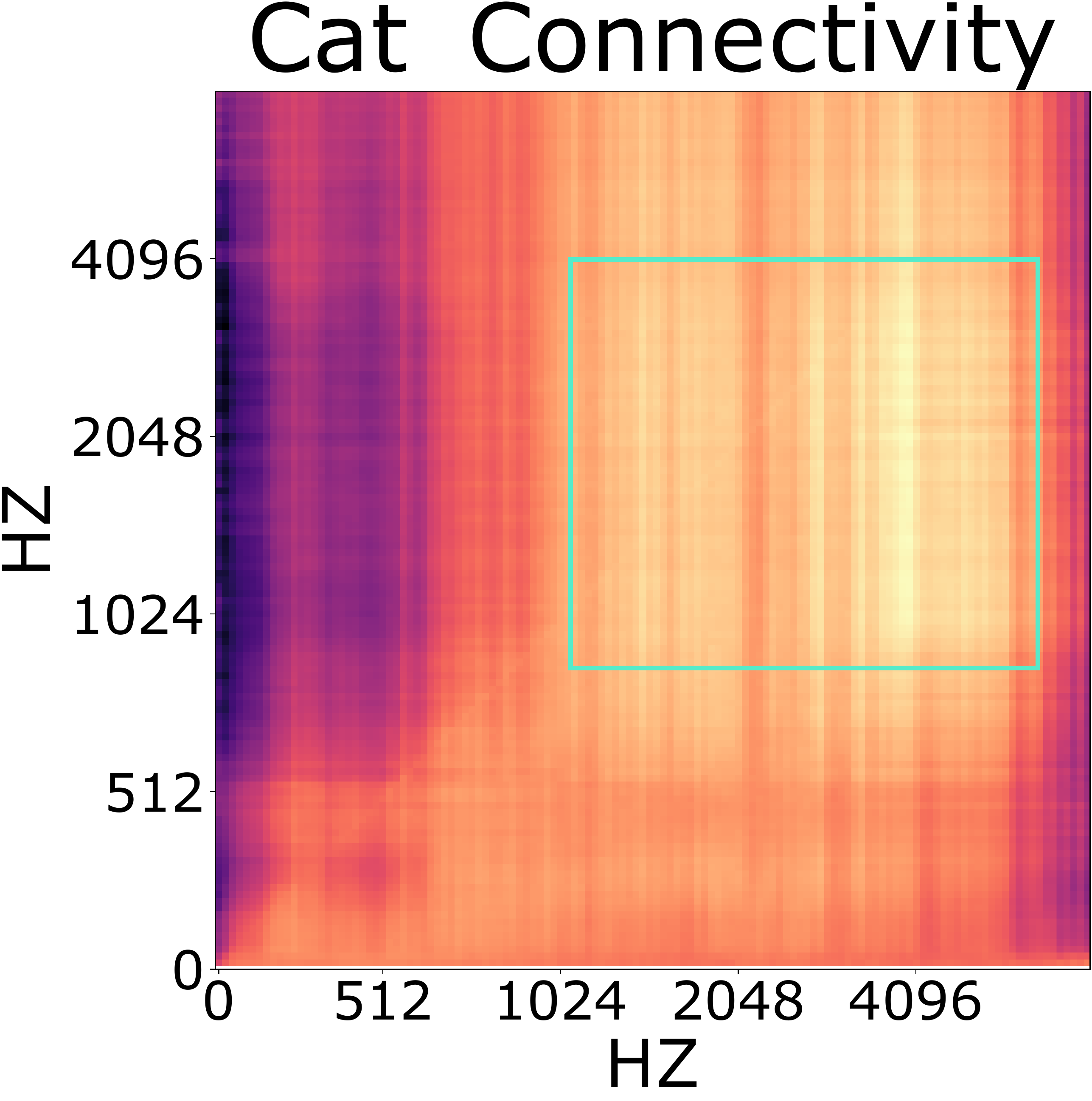}
\label{fig:Cat-ENC}
}
\subfloat[Nine EC]{
\includegraphics[height=0.475\columnwidth]{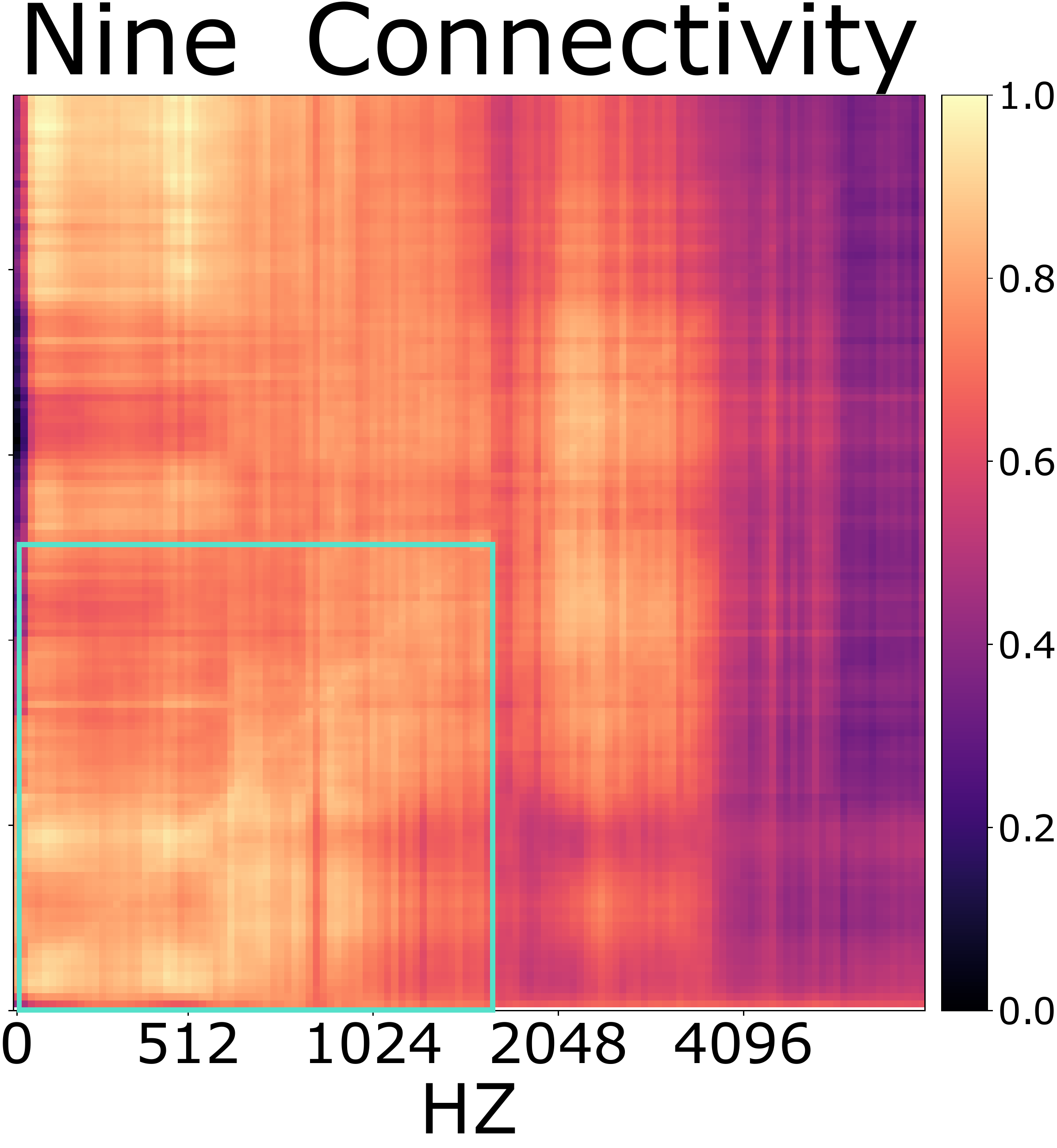}
\label{fig:Nine-ENC}
}
\caption{EC learned by our model for keywords 'cat' and 'nine' \textit{superimposed} with \textit{noise}. We used a test fold of $16$ subjects and computed mean EC with 10 trials per subject. Our model accurately gives high attention values to medium-to-high channels for 'cat' and low-to-medium channels for 'nine' samples.  \textbf{Average values:} Inside green box: $0.89$ for Fig. \ref{fig:Cat-ENC} and $0.78$ for Fig. \ref{fig:Nine-ENC}, outisde box: $0.69$ for Fig. \ref{fig:Cat-ENC} and $0.63$ for Fig. \ref{fig:Nine-ENC}. X and y axis denote the frequency channels in hertz (HZ).}
\label{fig:NLP ENC difference}  
\end{figure}

We first present the learned connectivity for the relatively easier task of NLP. We show the difference in the learned EC for the two keywords ('cat' and 'nine') in  Fig. \ref{fig:NLP ENC difference}. Fig.~\ref{fig:Cat-ENC} show high connectivity between higher channels, whereas Fig. \ref{fig:Nine-ENC} show high connectivity for relatively lower channels which follows the high frequency sounds in 'cat' and relatively lower frequency sounds in 'nine'. 


\begin{figure}[t!]
\centering
\subfloat[DECENNT ENC]{
\includegraphics[height=0.45\columnwidth]{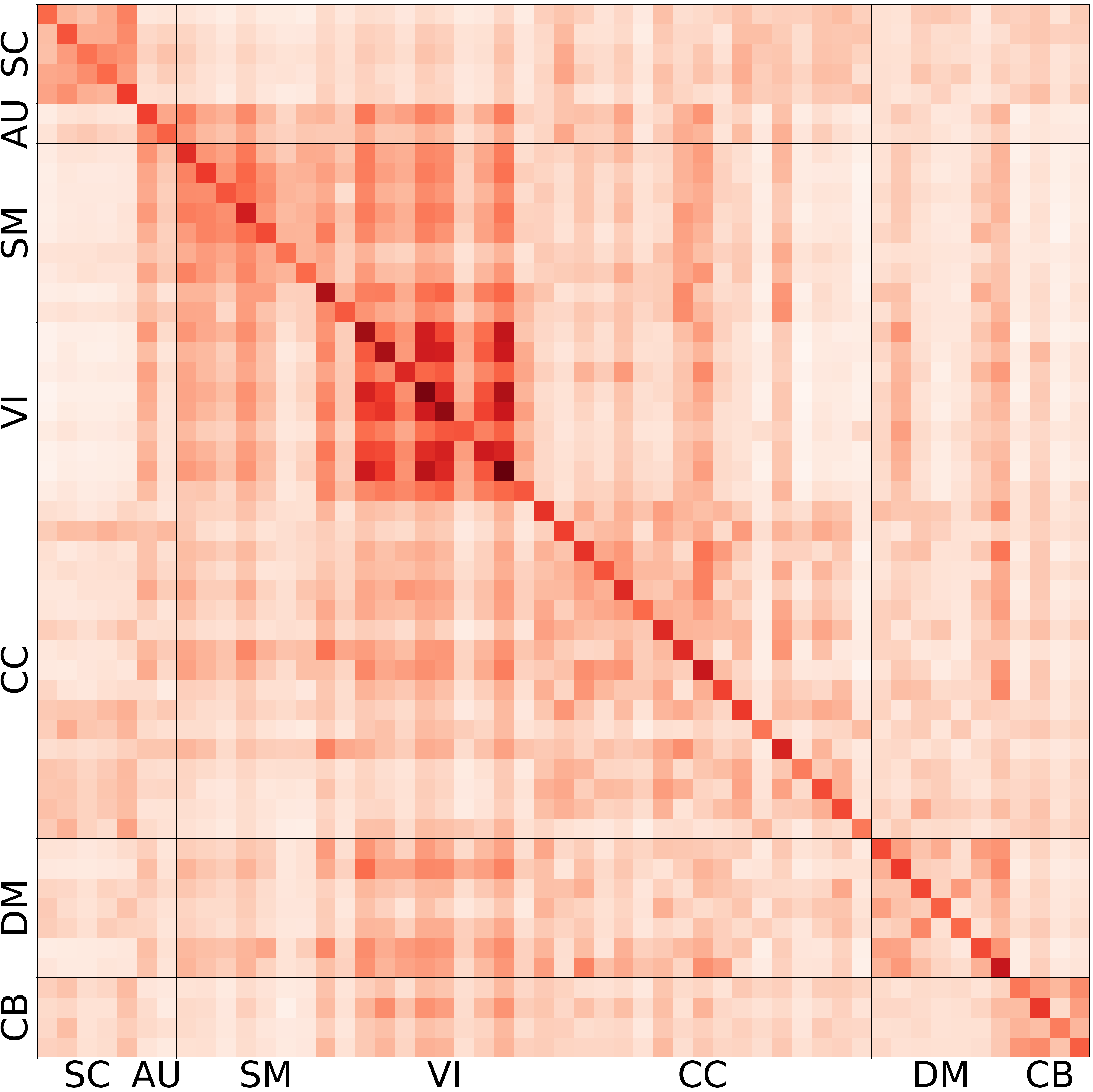}
\label{fig:TSA-ENC}
}
\subfloat[PCC FNC]{
\includegraphics[height=0.45\columnwidth]{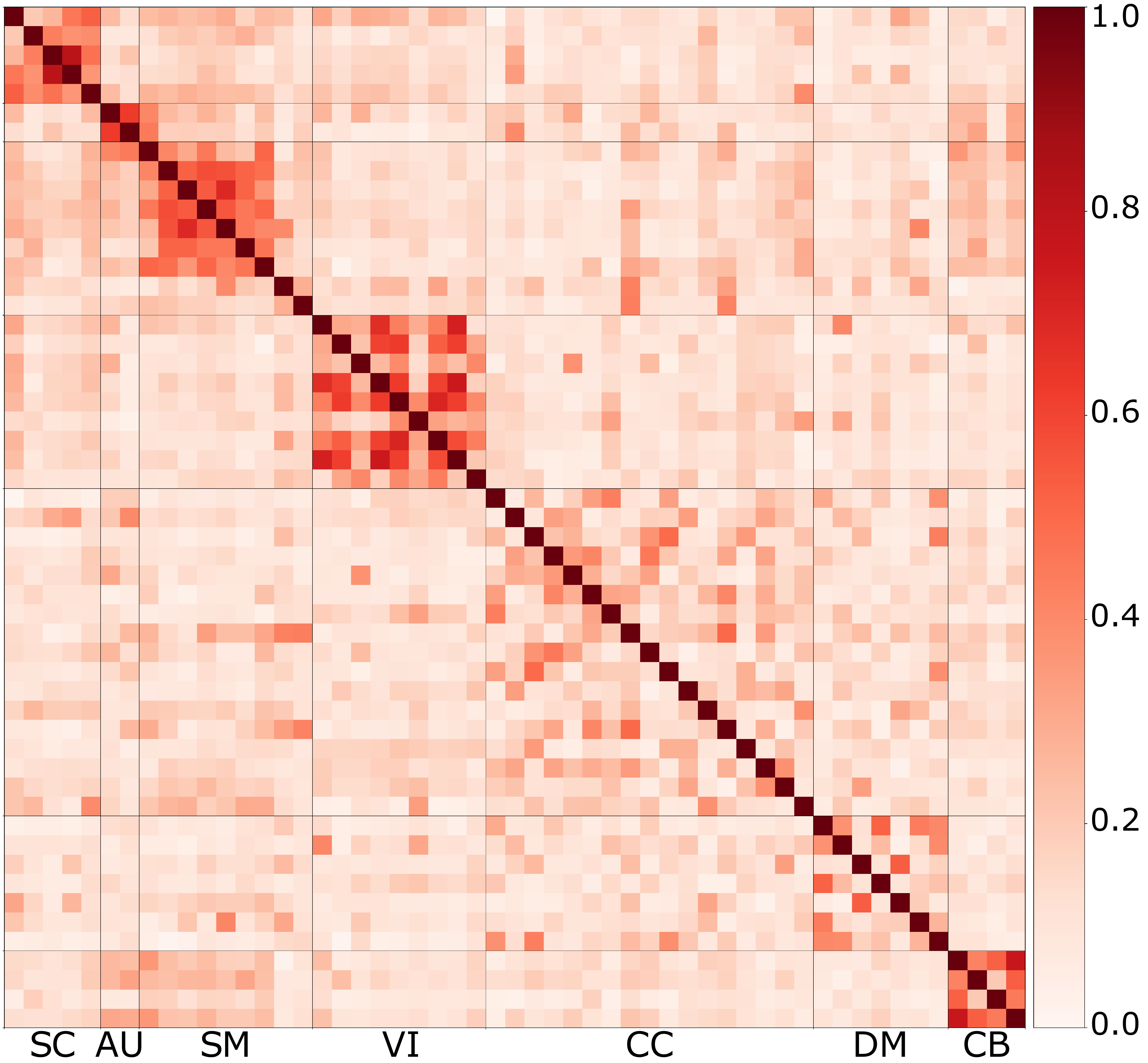}
\label{fig:PCC-ENC}
}
\caption{Fig. \ref{fig:TSA-ENC} is the connectivity matrix generated by our model for FBIRN dataset. We used a test fold of $16$ subjects ($10$ trials each) and computed mean ENC for all subjects. Fig. \ref{fig:PCC-ENC} is the mean FNC of the same subjects generated by PCC. Both figures are strikingly similar, which verifies the correctness of the connectivity matrix learned by our model. To match the positive weights of our model, we normalize the FNC from 0 to 1 instead of -1 to 1.}
\label{fig:FNC}
\end{figure}

Next we compare the connectivity matrix learned by our model on neuroimaging dataset with FNC computed PCC, which is probably the most popular method for computing connectivity matrix. Fig.~\ref{fig:FNC} shows that the two matrices are comparable, but our effective network connectivity (ENC) Fig.~\ref{fig:TSA-ENC} is directed and provides additional details. We also see that our ENC Fig.~\ref{fig:TSA-ENC} has more inter-network connectivity which is missing in Fig.~\ref{fig:PCC-ENC}. The effect of visual (VI) network onto other networks is seen only in Fig.~\ref{fig:TSA-ENC}. We group the ICA components according to \cite{10.3389/fnsys.2011.00002} into seven domains based on anatomical and functional properties. 53 components out of 100 fall into the seven domains and the rest are marked as noise. The connectivity matrix clearly shows that the components have high intra-domain connectivity, which matches the existing literature \cite{10.3389/fnsys.2011.00002}. 

\begin{figure}[ht!]
\centering
\includegraphics[width=\columnwidth]{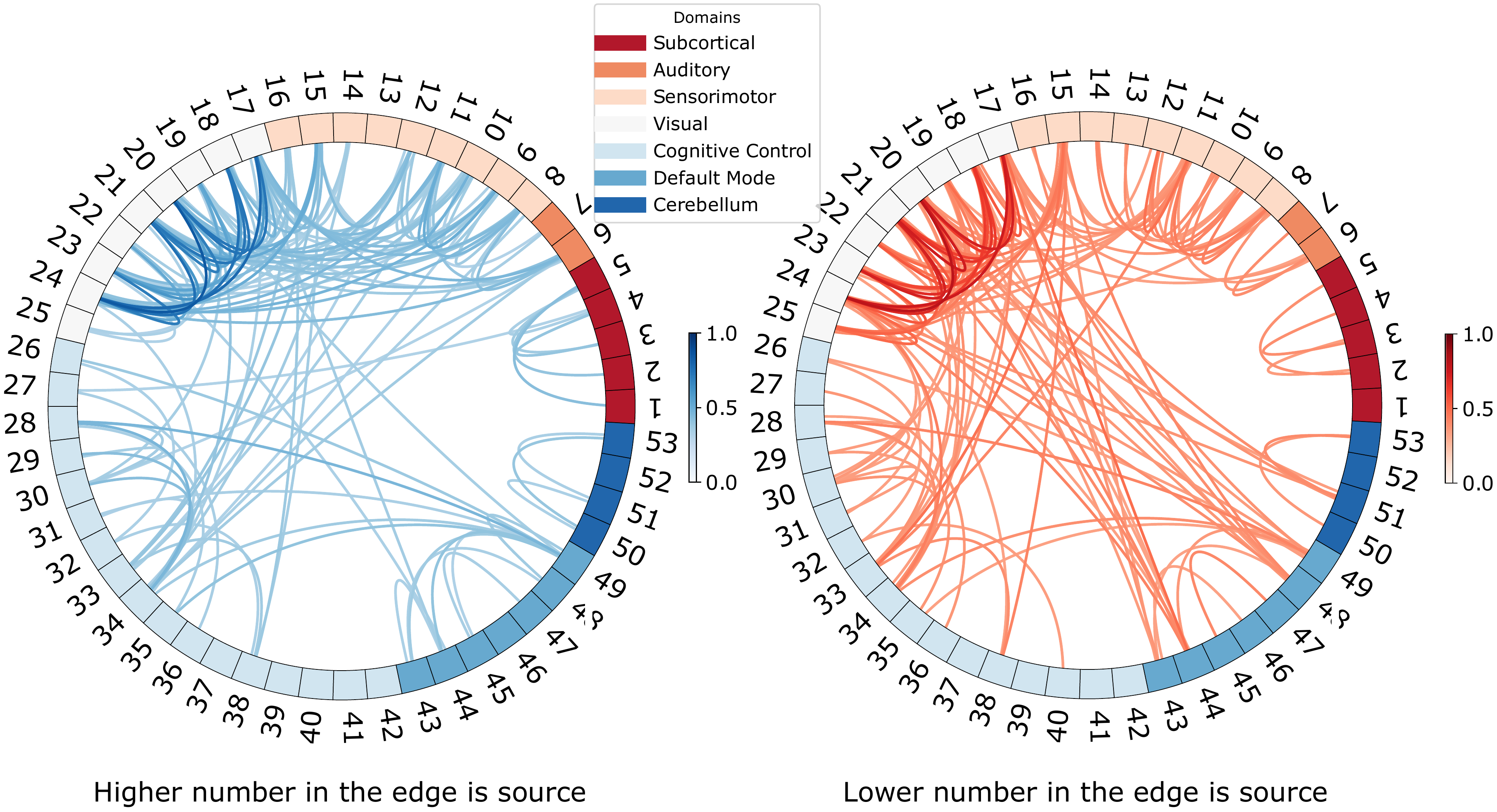}
\caption{Top 10$\%$ directed edges of FBIRN ENC \ref{fig:TSA-ENC}. The numbers represent the $53$ crucial components. The figure shows the direction of connectivity. Visual (VI) affects other domains, cognitive control affects sensorimotor. \textbf{Edges:} VI $\rightarrow$ other: 79, other $\rightarrow$ VI: 25. CC $\rightarrow$ SM: 9, SM $\rightarrow$ CC: 3. }
\label{fig:Connectome - FBIRN}
\end{figure}

Furthermore, as our model learns ENC, we use Fig. \ref{fig:Connectome - FBIRN} to show the importance of direction.  Fig. \ref{fig:Connectome - FBIRN} (left) shows edges from $a$ to $b$, where $a > b$ For example, the edge (8,21) means the edge is from $21$ to $8$. It is observable that the components in visual (VI) heavily affect components in sensorimotor (SM). The direction is reversed in  Fig. \ref{fig:Connectome - FBIRN} (right) and SM does not affect VI. Similar direction can be seen between cognitive control (CC) and SM. The presence of direction is of paramount importance and is missing from FNC. It can potentially help to make and answer interventions in data.

\subsection{Temporal Attention}

\begin{figure}[ht!]
\centering
\subfloat[All trials]{
\includegraphics[width=0.45\columnwidth]{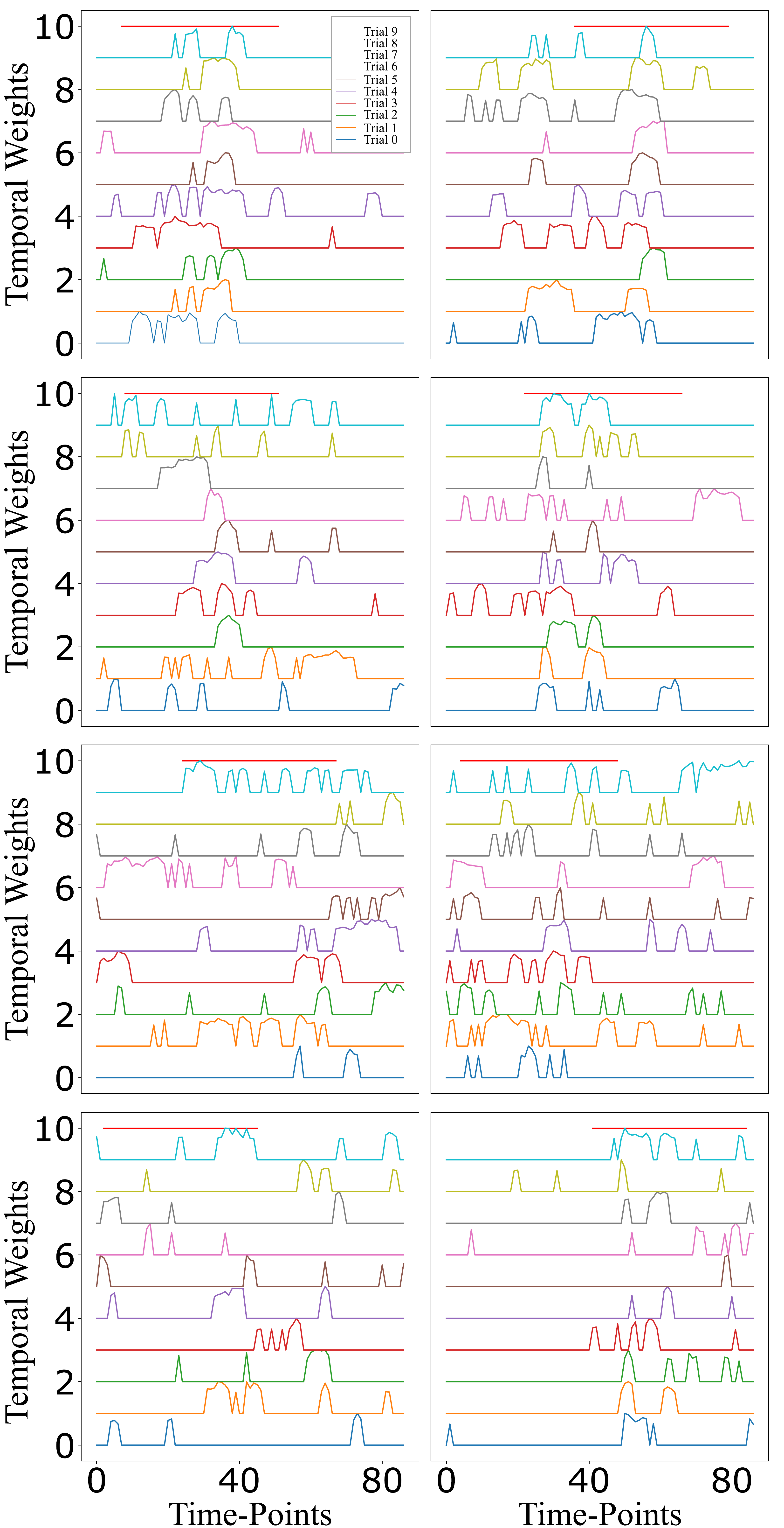}
\label{fig: Attention weights all}
}
\subfloat[Averaged]{
\includegraphics[width=0.45\columnwidth]{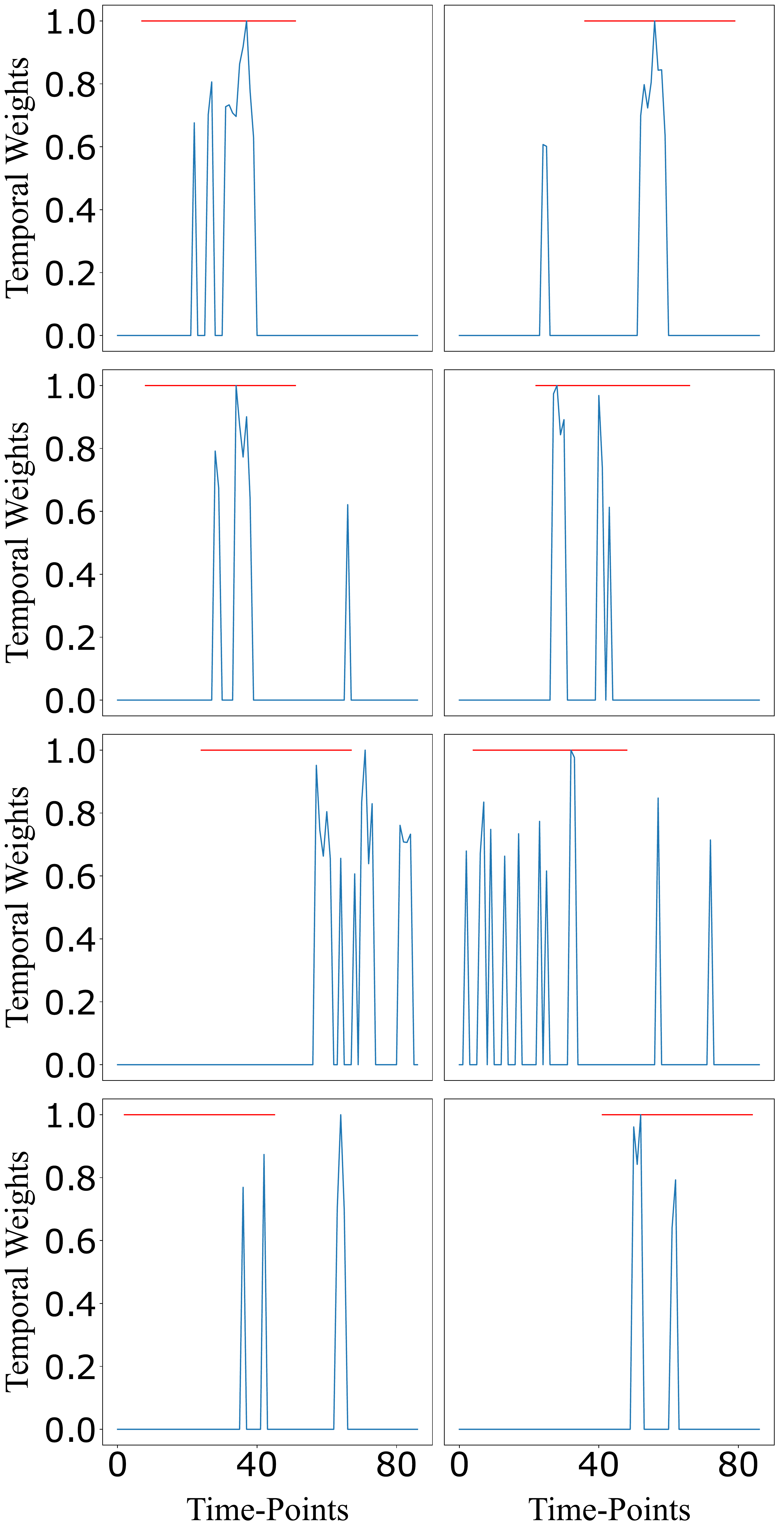}
\label{fig: Attention weights averaged}
}
\caption{Fig. \ref{fig: Attention weights all} is the normalized temporal attention weights for keyword detection task for $8$ subjects, with $10$ trials for each. To keep the lines separate, we added the trial number to the weights. Fig.~\ref{fig: Attention weights averaged} is the mean weights for the $10$ trials of the same subjects. The top red line marks the actual time-points for the keyword. \textbf{Statistics:} True Positive=56, False Positive=18, True Negative=331, False Negative=291, Precision=0.76, Sensitivity=0.16, Specificity=0.95.
}
\label{fig:Attention-weights-KW}
\end{figure}

As in rs-fMRI the subjects are not performing any specific task at any time-point, there is no available true knowledge of important time-points. Because of this reason, we show keyword detection experiments where the precise location of the keyword is available. Fig. \ref{fig:Attention-weights-KW} shows attention weights for 8 test subjects. The attended time-points match with the time-points of the keyword. This is extremely significant and proves the model can accurately find important time-points, as the location of the keyword was never given to the model. We compute the statistical values such as (precision, recall) of the temporal attention, mentioned in the caption of Fig.~\ref{fig:Attention-weights-KW}. We assign label '1' to time-points where 'cat' audio is superimposed and label '0' to all other time-points which gives us the true labels. For predicted labels, we assign label '1' to time-points with attention value greater than 0 and label '1' to all other time-points. The stats shows that the model a) assigns high attention values to 'cat' time-points, b) does not attend to 'non-cat' time-points and c) does not attend to all 'cat' time-points. Although, we would have liked the model to attend to all cat time-points, we think the model does not do that because of two reasons; 1) The 1-sec long 'cat' audio files on average have the 'cat' sound for only 0.5 seconds or less whereas, when creating  Fig.~\ref{fig:Attention-weights-KW} and the stats, we used the complete 1-sec time-points. 2) The model maybe looking for a part of the keyword 'cat' which is distinct from the noise.

 To further check the correctness of the time-points selected by our model and the affects on classification performance, we perform an experiment where after training the model, we compute $\W^f$ using only top $5\%$ attended time-points for training data to train an LR model and then use the top $5\%$ time-points for the test data to test the model. Similarly we perform experiments for bottom $5\%$ values as well. Tab. \ref{Table:top vs bottom 5 percent} shows the comparison using three brain datasets. The results show that the LR model provides high AUC score by just using top $5\%$ of the time-points attended by the model. Thus, it proves that a) not all time-points are important for classification of the downstream task and b) our model accurately finds the important time-points. We use an LR model for this experiment to show that the learned top/bottom $5\%$ time-points are not limited to our model but is generalized such that an independent LR module gives high classification results using the top $5\%$ attended time-points and does not learn on the low $5\%$ data. In our experiments, we also note upto $5\%$ drop in AUC when not using temporal-attention.

\begin{table}[ht]
\caption{AUC score comparison on brain datasets with ICA components by using all, top 5 \% and bottom 5\% time-points only.}
\centering
\begin{tabular}{@{}lllll@{}}
\toprule
            & Method  & FBIRN & OASIS & ABIDE \\ \midrule
100 \%      & DECENNT & 0.844 & 0.72  & 0.65  \\
Top 5 \%    & LR      & 0.835 & 0.713 & 0.642 \\
Bottom 5 \% & LR      & 0.566 & 0.548 & 0.532 \\ \bottomrule
\end{tabular}
\label{Table:top vs bottom 5 percent}
\end{table}


  


\section{Conclusion}
Our model demonstrates the importance of learning dynamic temporal graphs for any multivariate time series, which is currently missing from the existing literature. Using dynamic graphs, our model outperforms SOTA methods across five different tasks, proving that the model is applicable across different fields and tasks. By learning the correct graph structure/connectivity matrix for the data, our model eliminates the need for existing external methods such as PCC, K-means. Our model learns a directed graph structure that provides more detail than a symmetric correlation matrix which does not capture effective connectivity. As seen in results, our learned EC matrices give the direction of connectivity between brain regions. The temporal attention module proves to be highly effective in terms of classification. As shown in the paper, it provides stable attention weights and accurately finds the critical time-points depending on the downstream task. Both self and temporal attention modules result into stable, consistent attention values and increase the classification performance across tasks. These attributes address the questions regarding explainability of attention mentioned in \cite{jain2019attention,wiegreffe2019attention}.
Many tasks across many fields are ever dynamic and have missing graph structure, e.g. (Brain functional networks, social networks, self-driving cars etc.) which increases the need of methods like DECENNT. Temporal attention used in brain connectivity can help us find important bio-markers relative to the disorder/disease which in turn help us understand the disorder and its causes. For future work, we plan to extensively interpret the learned connectivity structures, and see the differences in them across controls and patients and across multiple brain disorders. We plan to compare our model with other methods of capturing brain's network connectivity such as transfer entropy and Granger causality. We also want to incorporate a form of spatial attention, which like temporal attention, could help identify essential nodes/components that are sometimes unavailable in many fields. We also for each class of subject want to examine how ENC changes overtime and if/how the direction of flow of information changes through time.

\section*{Acknowledgements}

This work was funded in part by NIH RF1MH121885, R01MH123610, R01EB006841 and NSF 2112455 grants. Data for healthy subjects was provided [in part] by the Human Connectome Project, WU-Minn Consortium (Principal Investigators: David Van Essen and Kamil Ugurbil; 1U54MH091657) funded by the 16 NIH Institutes and Centers that support the NIH Blueprint for Neuroscience Research; and by the McDonnell Center for Systems Neuroscience at Washington University. Data for Schizophrenia used in this study were downloaded from the Function BIRN Data Repository (\url{http://bdr.birncommunity.org:8080/BDR/}), supported by grants to the Function BIRN (U24-RR021992) Testbed funded by the National Center for Research Resources at the National Institutes of Health, U.S.A. Data for Alzheimer's was provided by OASIS-3: Principal Investigators: T. Benzinger, D. Marcus, J. Morris; NIH P50AG00561, P30NS09857781, P01AG026276, P01AG003991, R01AG043434, UL1TR000448, R01EB009352. AV-45 doses were provided by Avid Radiopharmaceuticals, a wholly owned subsidiary of Eli Lilly. Autism data was provided by ABIDE. We acknowledge primary support for the work by Adriana Di Martino provided by the (NIMH K23MH087770) and the Leon Levy Foundation and primary support for the work by Michael P. Milham and the INDI team was provided by gifts from Joseph P. Healy and the Stavros Niarchos Foundation to the Child Mind Institute, as well as by an NIMH award to MPM (NIMH R03MH096321).

\bibliography{IEEEabrv.bib,references.bib}{}
\bibliographystyle{IEEEtran.bst}


\end{document}